\documentclass{article} 
\usepackage{baichuan}

\usepackage{amsthm}
\usepackage{amsmath}
\usepackage{amssymb}
\usepackage{amsfonts}
\usepackage[utf8]{inputenc} 
\usepackage[T1]{fontenc}    
\usepackage{hyperref}       
\usepackage{url}            
\usepackage{booktabs}       
\usepackage{nicefrac}       
\usepackage{microtype}      
\usepackage{xcolor}         
\usepackage{multirow}
\usepackage{multicol}
\usepackage{graphicx}
\usepackage{subcaption}
\usepackage{utfsym}
\usepackage{svg}
\usepackage{array}
\usepackage{graphicx}
\usepackage{fancyhdr}
\usepackage{tablefootnote}
\usepackage{footnote}

\usepackage{makecell}
\usepackage{newtxmath}
\usepackage{bm}
\usepackage{bbding}
\usepackage{wrapfig}
\usepackage{placeins}
\newcommand{\tldt}{\textup{TLD}_t}
\newcommand{\llma}{\texttt{MetaGPT}}
\newcommand{\tr}{\text{tr}}

\newcommand{\vx}{\bm{x}}
\newcommand{\vh}{\bm{h}}

\newcommand{\vth}{\bm{\theta}}
\newcommand{\vt}{\bm{\tau}}

\theoremstyle{plain}
\newtheorem{theorem}{Theorem}

\newtheorem{property}[theorem]{Property}
\newtheorem{lemma}[theorem]{Lemma}

\theoremstyle{definition}
\newtheorem{definition}[theorem]{Definition}

\theoremstyle{remark}

\usepackage{colortbl}
\usepackage{xcolor}
\definecolor{mygray}{gray}{.85}
\newcommand{\eg}{\textit{e.g.},~}

\title{\centering MetaGPT: Merging Large Language Models \\
Using Model Exclusive Task Arithmetic}

\author{%
 	Yuyan Zhou$^{*}$ \\ 
	Baichuan Inc. \\
	\And
 Liang Song\thanks{Equal contribution} \\
	Baichuan Inc. \\
	\And
Bingning Wang\thanks{Corresponding author, \texttt{daniel@baichuan-inc.com}} \\  Baichuan Inc. 
	\And
	Weipeng Chen \\
	Baichuan Inc. 
}

\colmfinalcopy

\begin{document}

\maketitle

\begin{abstract}

The advent of large language models (LLMs) like GPT-4 has catalyzed the exploration of multi-task learning (MTL), in which a single model demonstrates proficiency across diverse tasks. 
Task arithmetic has emerged as a cost-effective approach for MTL. 
It enables performance enhancement across multiple tasks by adding their corresponding task vectors to a pre-trained model.
However, the current lack of a method that can simultaneously achieve optimal performance, computational efficiency, and data privacy limits their application to LLMs.
In this paper, we propose \textbf{M}odel \textbf{E}xclusive \textbf{T}ask \textbf{A}rithmetic for merging \textbf{GPT}-scale models (\llma), which formalizes the objective of model merging into a multi-task learning framework, aiming to minimize the average loss difference between the merged model and each individual task model. 
Since data privacy limits the use of multi-task training data, we leverage LLMs' local linearity and task vectors' orthogonality to separate the data term and scaling coefficients term and derive a model-exclusive task arithmetic method.

Our proposed \llma~is data-agnostic and bypasses the heavy search process, making it cost-effective and easy to implement for LLMs.
Extensive experiments demonstrate that \llma~leads to improvements in task arithmetic and achieves state-of-the-art performance on multiple tasks.

\end{abstract}

\section{Introduction}
\begin{figure}
    \centering
    \vspace{3mm}
    \includegraphics[width=0.7\textwidth]{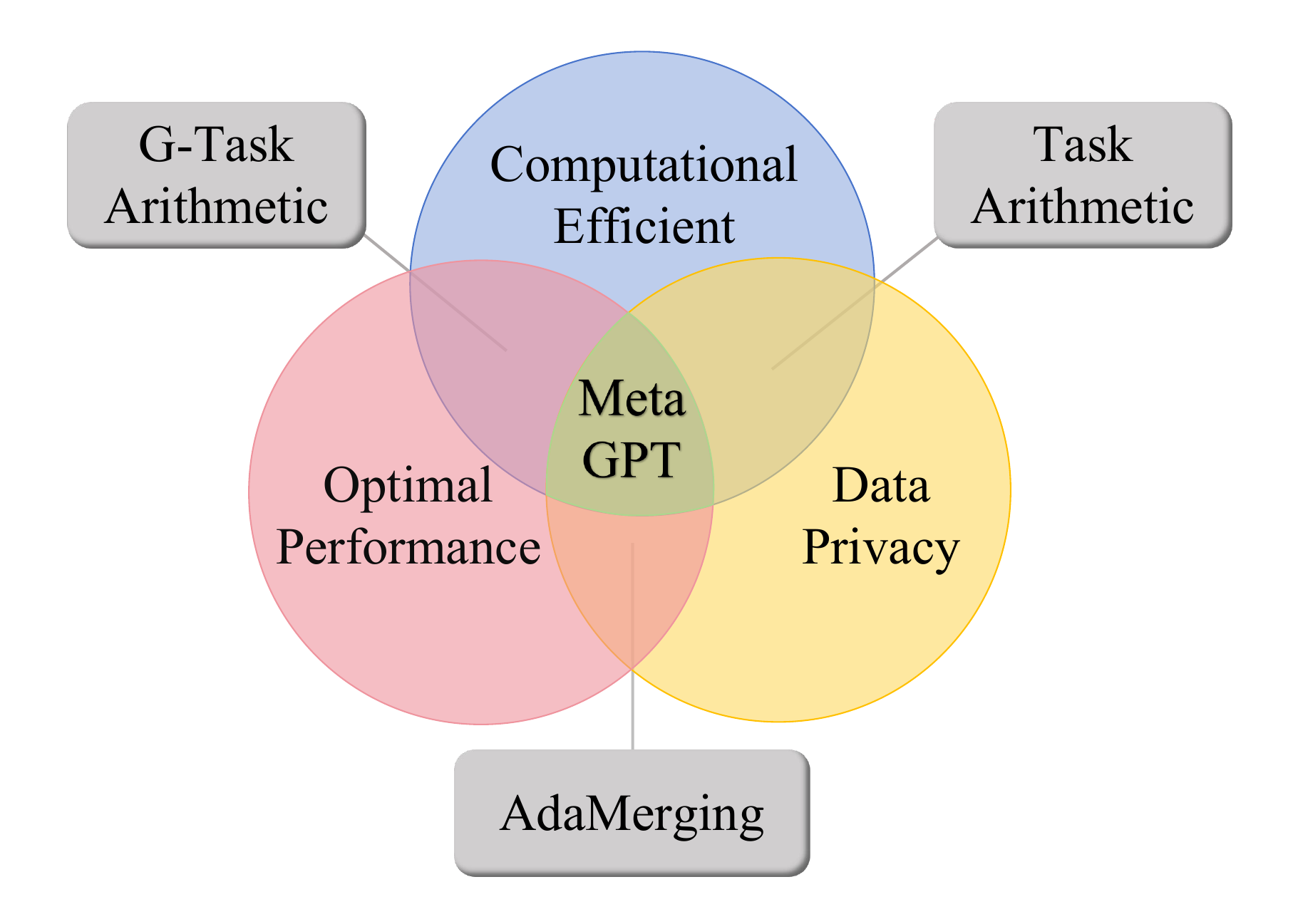}
    \caption{Existing methods face the trilemma of performance, data privacy, and computational costs, which hinders its application to LLMs. Our \llma~can solve these problems under careful approximation and thus can scale to GPT3-scale LLMs.}
    \label{fig:main-fig}
\end{figure}
In recent years, a well-established paradigm for AI has been to pre-train models using large-scale datasets and then to fine-tune the models on different tasks through supervised learning with task-specific datasets, which can lead to improved performance while requiring less labeled data~\cite{devlin2018bert,gpt4,dodge2020fine,baichuan2}.
However, for each new application, a separate model has to be fine-tuned and deployed, which is computationally expensive and resource-intensive~\cite{fifty2021efficiently, zhang2021survey}. Thus, Multi-Task Learning (MTL) methods have been proposed and developed to enable a single model to solve multiple tasks concurrently.

Conventional MTL approaches typically involve collecting raw data across multiple tasks and then jointly training a single model \cite{caruana1997multitask,yang2023adatask}. However, the fine-tuning process becomes extremely computationally intensive with the development of large language models (LLMs) that may comprise billions or even trillions of parameters. Therefore, researchers have explored merging various task-specific models with the expectation that the merged model can handle multiple tasks simultaneously.

One of the outstanding merging methods is task arithmetic~\cite{ilharco2022editing}. For a given task, the element-wise difference between the weights of the pre-trained model and the fine-tuned model is referred to as the task vector. Recent studies have shown that linearly adding multiple scaled task vectors to the pre-trained model can improve performance across those tasks \cite{ilharco2022editing,yang2023adamerging}. Nevertheless, previous task arithmetic methods face a trilemma in practice. 1) The best-performing task arithmetic methods require extra training to obtain optimal hyper-parameters, but the high computational costs hinder their application to GPT3-scale LLMs. 2) Some training-free methods heuristically set the scaling coefficient to a constant (\eg 0.3), which is efficient but leads to sub-optimal performance. 3) Some methods conduct grid search on the training/validation set, which is sometimes impractical and faces the risk of data privacy concerns.
In summary, as illustrated in Figure \ref{fig:main-fig}, there is essentially no task arithmetic method suitable for billion-scale models that perform satisfactorily in practice.

To address the aforementioned problems, in this paper, we propose \texttt{MetaGPT}: an \textit{optimal} and \textit{efficient} task arithmetic method for MTL \textit{without any data} (\textbf{m}odel \textbf{e}xclusive \textbf{t}ask \textbf{a}rithmetic).
We begin by providing a detailed theoretical analysis of the task loss difference and average loss difference introduced by the task arithmetic algorithm.
Since we aim to choose parameters that minimize the average loss difference, we first separate the data term and scaling coefficients, which also establishes a performance upper bound for task arithmetic. After separating the scaling coefficients, the final result is quadratic for each scaling coefficient, leading to a closed-form solution that is simple and effective to implement.

The experimental results on the LLaMA-2 \cite{touvron2023LLaMa} and Mistral \cite{jiang2023mistral} series demonstrate that the \llma~approach is superior to previous merging methods on several tasks. \llma~provides an efficient avenue to optimally implement task arithmetic for large-scale multi-task learning (MTL) and push the frontiers of language model merging. To sum up, our contributions include:
\begin{enumerate}
    \item We provide the mathematical formulation of the optimization objective for task arithmetic and the first theoretical analysis of the performance bound for task arithmetic.
    \item To achieve efficient, optimal, and model-exclusive task arithmetic, we separate the data term and scaling coefficients in the optimization objective, which leads to a closed-form solution for the scaling coefficients.
    \item Our \llma~is orthogonal to existing task vector-improving methods and can be integrated to achieve higher performance.
    \item Extensive experiments demonstrate that our \llma~can improve task arithmetic and achieve state-of-the-art performance.
\end{enumerate}

\section{Related Work}

\noindent\textbf{Model Merging.}
Currently, model merging has been developed for multiple uses such as improving performance on a single target task~\cite{izmailov2018averaging, wortsman2022model, zheng2024weak},  improving out-of-domain generalization~\cite{rame2023model, cha2021swad, arpit2022ensemble}, and improving the performance of multi-task learning~\cite{ilharco2022editing, yadav2024ties, yu2023language, huang2024emr, ye2023merging}, which is the core focus of our research.
The range of applications has led to a proliferation
of methods to improve beyond simple parameter averaging.
Fisher merging~\cite{matena2022merging} tries to weight the importance of individual models using
Fisher Information Matrix and uses it to merge different models.
RegMean~\cite{jin2022dataless} formulate the merging problem as a regression problem and leads to an optimal solution for linear models.
Task Arithmetic~\cite{ilharco2022editing} presents a method for merging models by adding task vectors to the pre-trained model to improve multi-task performance.
Ties Merging~\cite{yadav2024ties} and DARE~\cite{yu2023language}
propose to refine the task vectors by resolving the interference and removing extremely redundant components.
\citet{ortiz2024task} propose that fine-tuning the models in their tangent space can amplify weight disentanglement and lead to substantial performance improvements.

\noindent\textbf{Multi-Task Learning.}
Multi-task learning is a powerful method for solving multiple correlated tasks simultaneously~\cite{caruana1997multitask}. 
Current MTL works mainly focus on learning the shared representations from designing specific architecture~\cite{Misra2016CrossStitch, Sun2020Adashare} or using specific optimization methods~\citep{Sener2018Multi,liu2021conflict}.
The former focuses on learning the shared representation using different methods such as designing specific representation sharing module~\citep{liu2019end,ding2021MSSM}, learning to branch~\citep{lu2017fully,guo2020learning}, and based selection criteria~\citep{ma2018modeling,Hazimeh2021dselects}. 
And the latter focuses on balancing multiple tasks from the perspectives
of task training weights~~\citep{Sener2018Multi,liu2019end}, gradient dominance~\citep{zhao2018gradnorm,he2022MetaBalance,yang2023adatask}, and solving gradient conflicts~~\citep{yu2020gradient,chen2020just,liu2021conflict}. 
However, the conventional MTL approaches for collecting raw data across multiple tasks for joint training are not suitable for LLMs.
The factors contributing to this issue are twofold: first, computational inefficiency due to the substantial computational costs associated with updating pre-trained models; second, a significant number of data proprietors are reluctant to disclose valuable or privacy-sensitive raw data. 
\begin{figure*}
   \hspace{-.9cm}
    \includegraphics[width=1.1\textwidth]{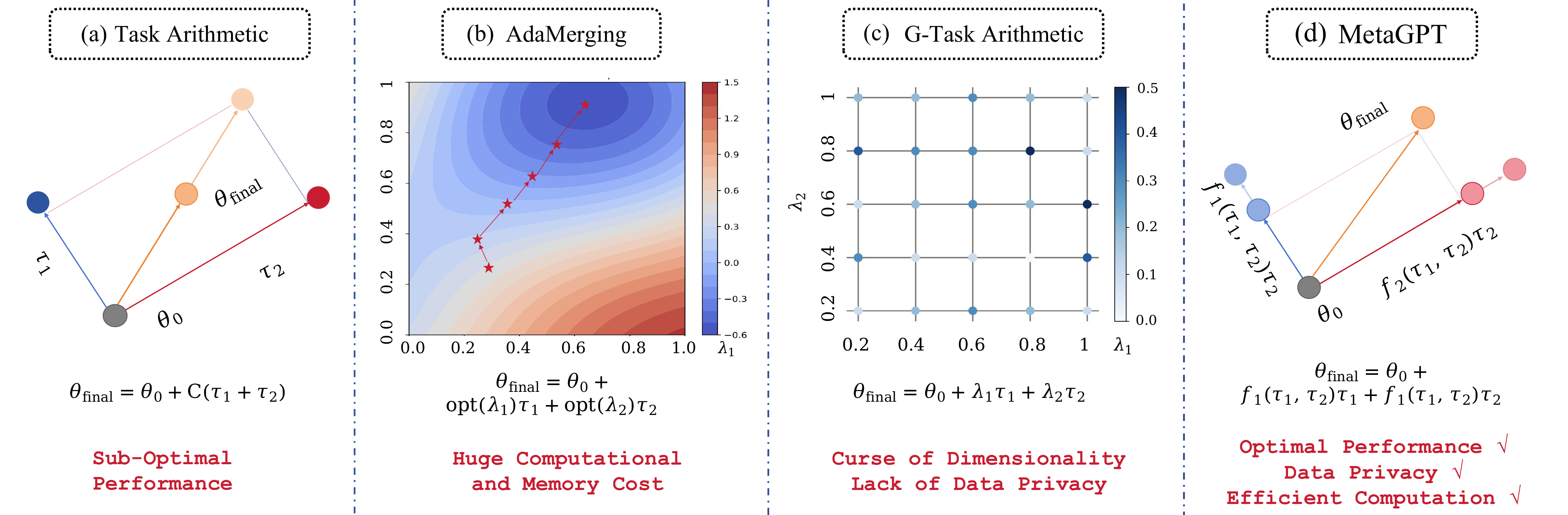}
    \caption{Current task arithmetic based methods face the problems of sub-optimal
performance, huge computational and memory cost, curse of dimensionality and data privacy, which makes it difficult to scale to LLMs. Our method solves the aforementioned problems and provides an avenue to scale task arithmetic to LLMs.}
    \label{fig:enter-label}
\end{figure*}
\section{Preliminaries}
\subsection{Notation} 
Let $f:\mathcal{X}\times\Theta\to\mathcal{Y}$ be a neural network taking inputs $\vx\in\mathcal{X}$ and parameterized by a set of weights $\vth \in\Theta$.
We assume $\mathcal{X} \subseteq \mathbb{R}^{p}$, $\Theta \subseteq\mathbb{R}^m$ and $\mathcal{Y}\subseteq\mathbb{R}^q$.
We consider fine-tuning a pre-trained model $f(\cdot,\vth_{0})$ on $T$ different tasks, with each task $t$ consisting of a triplet $( \mathcal{D}_t, \mathcal{L}_t, \vth_t)$ 
, where $\mathcal{D}_t = (\mathcal{D}^{\text{train}}_t, \mathcal{D}^{\text{val}}_t, \mathcal{D}^{\text{test}}_t)$ is the training, validation and test data of task $t$, 
$\mathcal{L}_t$ is the loss function of task $t$, and $\vth_t$ is the model parameters fine-tuned on task $t$ based on the pre-trained weight $\vth_0$. 

\subsection{Task Arithmetic}  
Let the \textit{task vector} of task $t$ be the difference between the fine-tuned and the pre-trained weights: 
\begin{equation}
    \vt_t = \vth_t - \vth_0.
\end{equation}
Task arithmetic aims to solve the multi-task learning problem by directly adding the scaled task vectors to the pre-trained model weight $\vth_0$:
\begin{equation}
    \vth_{\text{final}} = \vth_{0} +\sum_{i=1}^T\lambda_i \vt_i
    \label{taeq}
\end{equation}
where $\lambda_i$ is the scaling coefficient of task vector $\tau_i$. As illustrated in Eq.~\ref{taeq}, the task arithmetic introduces $T$ hyper-parameters $\{\lambda_i | i=1,\cdots, T\}$ and the choice of these scaling coefficients has a significant influence on the performance of the merged model. 
Thus, selecting the appropriate scaling coefficients for different task vectors remains a challenging problem.

\subsection{Existing Methods}
Earlier task arithmetic~\cite{ilharco2022editing, yadav2024ties} propose to perform a grid search (G-Task Arithmetic) on the validation set to choose the optimal scaling coefficients.
However, as the number of tasks increases, exploring all the scaling coefficient combinations faces the curse of dimensionality. 
Therefore, to simplify the problem, they use the same value for multiple scaling coefficients, thereby reducing the computational complexity.
In the absence of the training/validation data, they set $\lambda=0.3$ as the default setting for dataless arithmetic.
Moreover, Adamerging~\cite{yang2023adamerging} aims to autonomously learn the coefficients from unlabeled test samples using entropy minimization.

\subsection{Scalability Challenges for LLMs}
The methods mentioned above are not suitable for scaling to LLMs:
The grid search method requires extra validation/training data, which faces the risk of data privacy concerns and the curse of dimensionality when the number of tasks increases.
For instance, conducting a grid search for three hyper-parameters, each with a discretization interval of 0.01, would require $10^6$ forward passes across the entire dataset.
Setting a fixed value such as $0.3$ for all the $\lambda_i$ is time-efficient and can be applied to LLMs, but it leads to sub-optimal performance.
Using test data input to unsupervised optimize these hyper-parameters can lead to an optimal solution but requires extra data and necessitates loading multiple models for training.
This process is both time and memory consuming, making it challenging to apply to LLMs. 
For example, merging three LLMs requires loading three LLMs simultaneously to optimize, which is extremely costly.
The statement above suggests that scaling up existing optimal task arithmetic to LLMs remains a challenging problem. 
\section{Our Proposed \llma}
\label{method}
\subsection{Overview} 
To solve the problems above, we propose a new algorithm \llma, based on careful approximations to a closed-form solution, which easily scales to giant models both in terms of runtime as well as performance while protecting data privacy.
In this section, we state the motivation and optimization problem and solve it step by step.
All proofs of lemmas and theorems are provided in the appendix.

\subsection{\llma~Optimization Objective}
\begin{definition}[Single Task Loss Difference]
    For the fine-tuned model $\vth_i$ and the task arithmetic merged model $\vth_{\text{final}}$. The Task Loss Difference in task $t$ ($\text{TLD}_t$) is defined as:
    \begin{align}
    \label{tldeq}
        \textup{TLD}_{t}(\lambda_1, \cdots, &\lambda_T, \vt_1, \cdots, \vt_T) =  \mathcal{L}_t(\vth_{\textup{final}}, \vx) - \mathcal{L}_t(\vth_{t}, \vx).
    \end{align}
\end{definition}
It is obvious that smaller $\tldt$ suggests that the loss of the merged model is close or even lower than the fine-tuned model on task $t$, which indicates a better task arithmetic performance. 

However, for task arithmetic, it aims to improve the average performance of the final model on all the tasks. 
Thus, we define the average of all the task loss differences as Average Loss Difference (ALD), which can be formulated as follows:
\begin{definition}[Average Task Loss Difference]
    For the fine-tuned models $\{\vth_i | i=1, \cdots, T\}$ and task arithmetic merged model $\vth_{\text{final}}$. The average loss difference for all tasks is defined as:
    \begin{align}
    \label{aldeq} 
        \textup{ALD}(\lambda_1,& \cdots, \lambda_T, \vt_1, \cdots, \vt_T) = \frac{1}{T}\sum_{t=1}^T\left( \mathcal{L}_t(\vth_{\textup{final}}, \vx) - \mathcal{L}_t(\vth_{t}, \vx)\right).
    \end{align}
\end{definition}
Thus, the optimization objective of \llma~is to find the optimal scaling coefficients that can minimize the ALD, which can be formulated as:
\begin{definition}[Optimization objective of \llma]
Our \llma~aims at finding the scaling coefficients $\{\lambda_i|i=1,\cdots, T\}$, which minimizes the average loss difference ALD:
\begin{align}
    \mathop{\arg\min}_{\lambda_1, \cdots, \lambda_T}  \frac{1}{T}\sum_{t=1}^T\left( \mathcal{L}_t(\vth_{\textup{final}}, \vx) - \mathcal{L}_t(\vth_{t}, \vx)\right).
    \label{optimization objective eq}
\end{align}
\end{definition}
\subsection{Separating Data and Coefficients}
Before analyzing ALD, we start with reformulating $\tldt$ by its Taylor expansion.
\begin{lemma}
\label{LLema4}
    Using Taylor expansion for $\mathcal{L}(\vth_{\textup{final}}, \bm{x})$ at $\vth_t$, the $\textup{TLD}_t$ in Eq.~\ref{tldeq} can be reformulated as a quadratic form with respect to the linear combination of $\bm{\lambda}$ and $\vth$: 
    \begin{align}
        \textup{TLD}_t 
        &=\frac{1}{2}\bm{h}_t^\top\left(\int_{0}^{1}\nabla^2 \mathcal{L}_t(\gamma_t(\beta))d\beta \right)\bm{h}_t,
        \label{tldhessian}
    \end{align}
    where $\gamma_t(\beta) = \vth_t+\beta(\vth_{\textup{final}} - \vth_t)$ and $\bm{h}_t$ is the linear combination of $\bm{\lambda}$ and $\vth$:
    \begin{align}
    \label{tldeq2}
        \bm{h}_t = \sum_{k\neq t} \lambda_k(\vth_k - \vth_0) - (1- \lambda_{t})(\vth_{t} - \vth_0).
    \end{align}
\end{lemma}
Single $\tldt$ is associated with the data, models, and scaling coefficients.  As we can see in Eq.~\ref{tldhessian}, we have transformed the data term $\vx_t$ to the Hessian, the coefficients $\bm{\lambda} = [\lambda_1, \cdots, \lambda_T]$ and models term $[\vth_1, \cdots, \vth_T]$ to $\bm{h}$.
As our method tends to achieve model-exclusive task arithmetic, the final result should not correlate with the data term.
Thus, we first provide a property, which will be used latter in our theorem proofs to separate the data term and scaling coefficients and models term.
In general, if a pre-trained network $f(\cdot ; \vth_0)$ demonstrates kernel behavior during fine-tuning, i.e., fine-tuning occurs in the linear regime, the following property must be satisfied~\cite{jacot2018neural}:
\begin{property}[NTK linearization] 
\label{linear}
Around the initialization weights $\vth_0$, a neural network can be approximated with a linear approximation:
\begin{align}
f(\vx;\vth_0+\alpha(\vth_t - \vth_0))\approx f(\vx;\vth_0)+\alpha\cdot C .\label{eq:linearization}
\end{align}
where $C = (\vth_t - \vth_0)^\top \nabla f(\vx , \vth_0)$ is a data and model dependent constant.
\end{property}
It is worth noting that, as the network width approaches infinity, Eq.~\ref{eq:linearization} becomes exact and remains valid throughout training~\cite{jacot2018neural,arora2019exact, lee2019wide}, which is specifically suitable for the LLMs arithmetic scenario.

The second property is observed by \citep{ilharco2022editing}, which states that the different task vectors are orthogonal:
\begin{property}[Orthogonality of Task Vectors]
\label{orthogonality}
For task vector $\vt_i = \vth_i - \vth_0$ and $\vt_j = \vth_j - \vth_0$ ($i\neq j$), we have the following equation: 
\begin{equation}
    \vt_i^\top \vt_j = (\vth_i - \vth_0)^\top(\vth_j - \vth_0) = 0.
    \label{eq.orthogonal}
\end{equation}
\end{property}
Now, as we previously introduce our first Lemma to transform the $\tldt$ in Eq.~\ref{tldeq} into a quadratic form with respect to the linear combination of $\bm{\lambda}$ and $\vth$.
Next, using Property~\ref{linear},\ref{orthogonality} and Lemma~\ref{tldeq2}, we can upper bound the $\tldt$ and separate the data term and scaling coefficients and models term. 
\begin{theorem}
\label{thm7}
    The $\tldt$ can be upper bounded by:
    \begin{align}
        &\tldt(\lambda_1, \cdots, \lambda_T, \vt_1, \cdots, \vt_T) \leq 
         \frac{\delta_t^2}{2} \left\Vert \vth_t - \vth_0  \right\Vert_2^2 \bigg\{ \sum_{k\neq t}^T
        \vmathbb{1}_t(\lambda_k^2)
        \Vert \vth_k -  \vth_0\Vert^2 \bigg\},
    \end{align}
    \textup{where} $\delta_t$ is a data-dependent constant and we use $\vmathbb{1}_t(\lambda_k^2)$ to denote $(\lambda^2_k)\vmathbb{1}(k\neq t) +(1 - \lambda^2_k)\vmathbb{1}(k= t)$.
\end{theorem}
Now, after separating the data-related term to $\delta_t$, the scaling coefficients and models term to $\vmathbb{1}_t(\lambda_k^2)$. By summing all the $\tldt$s, we can separate the two terms for ALD: 
\begin{theorem}
\label{thm8}
    By summing all the $\tldt$, we can separate the correlation between data term and scaling coefficients term in \textup{ALD}:
    \begin{align}
        \label{ald decoupled}
        \textup{ALD}&(\lambda_1, \cdots, \lambda_T, \vt_1, \cdots, \vt_T) \leq \sum_{t=1}^T \delta_t^2 \left\Vert \vth_t - \vth_0  \right\Vert_2^2 \left\{\sum_{k\neq t}^T
        \vmathbb{1}(\lambda_k^2)
        \Vert \vth_k -  \vth_0\Vert^2 \right\},
    \end{align}
\end{theorem}
\subsection{The Optimal Solution}
After separating the data term and the scaling coefficients term, we can now reformulate our optimization objective Eq.~\ref{ald decoupled} and derive the closed-form optimal solution of the scaling coefficients.
\begin{theorem}[$\lambda$ decomposition of ALD]
\label{thm9}
    For each $\lambda_t$, we use it to decompose Eq.~\ref{ald decoupled} as:
    \begin{align}
        \textup{ALD}\leq 
        \sum_{t=1}^T \textup{ALD}_{\lambda_t},
    \end{align}
    where $\textup{ALD}_{\lambda_t}$ is:
    \begin{align}
    \label{single_decouple}
        \textup{ALD}_{\lambda_t} =&\frac{\delta_0^2}{2} \Vert \vth_t  -\vth_0\Vert^2 \left[\sum_{k=1}^T \vmathbb{1}_t(\lambda) \Vert \vth_k - \vth_0\Vert^2\right],
    \end{align}
\end{theorem}
where $\delta_0 = \max_{t}\delta_t$. The equation above easily leads to a model-exclusive closed-form solution:
\begin{theorem}[Optimal Scaling Coefficients]
\label{thm10}
    We can solve $\lambda_t$ form Eq~\ref{single_decouple} by: 
    \begin{align}
        \lambda_t = \mathop{\arg\min}_{\lambda_t} \Vert \vth_t - \vth_0 \Vert^2\left[\sum_{k=1}^T \vmathbb{1}_t(\lambda) \Vert \vth_k - \vth_0\Vert^2\right].
    \end{align}
    The above equation is quadratic on $\lambda_t$ and the optimal solution for $\lambda_t$ is:
    \begin{align}
        \lambda_t = \frac{\Vert \vth_t - \vth_0\Vert^2}{\sum_{k=1}^n\Vert \vth_k - \vth_0\Vert^2}.
        \label{solution}
    \end{align}
\end{theorem}

\begin{figure}[htb]
    \centering
    \begin{minipage}[b]{0.45\textwidth}
        \centering
        \includegraphics[width=\textwidth]{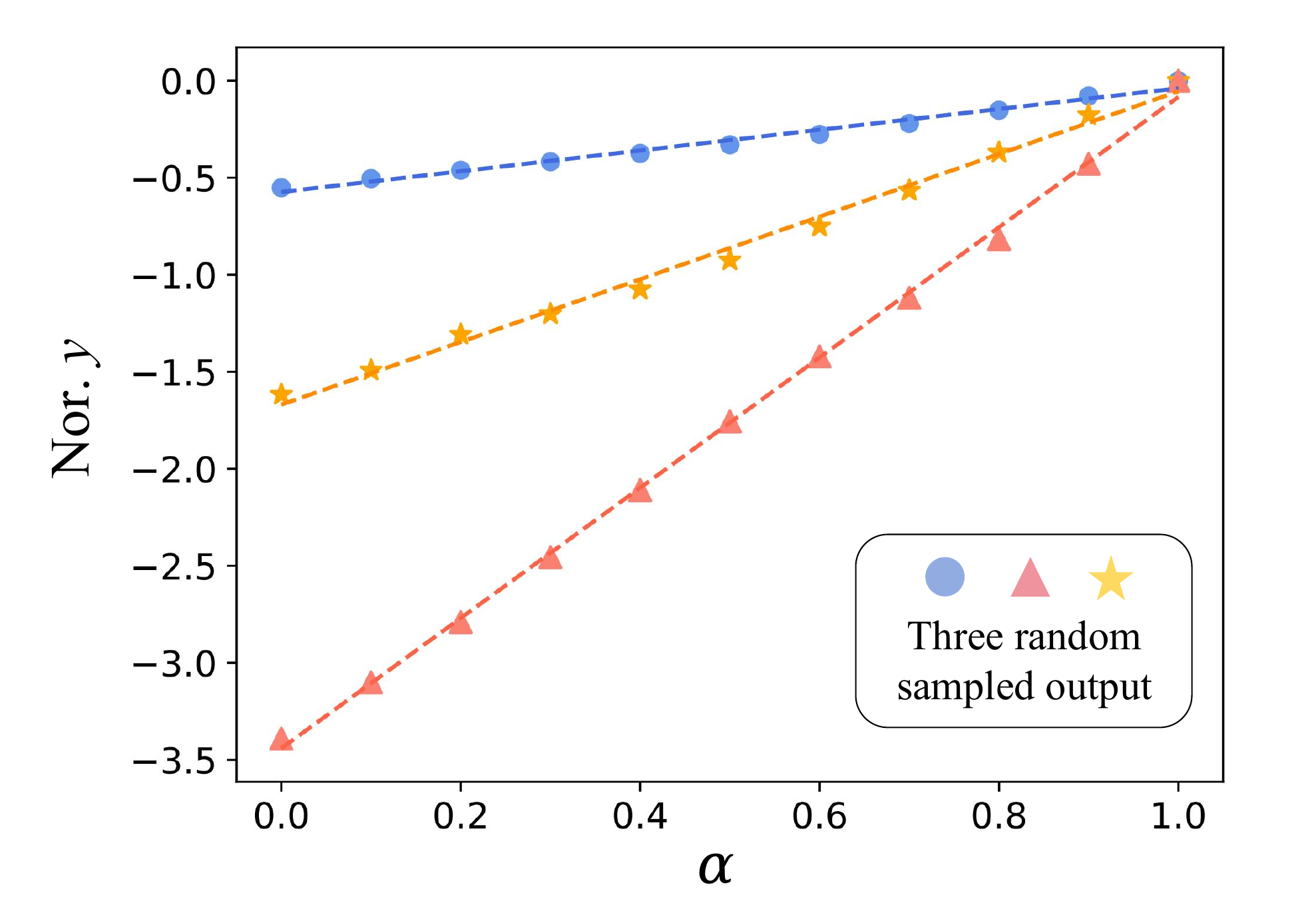}
        \caption{Verification of NTK linearization. We randomly sampled the outputs of Llama-2-7b-chat-hf with different $\alpha$. We can see that the sampled outputs are linearly with $\alpha$ as expected.}
        \label{linear_fig}
    \end{minipage}
    \hspace{0.08\textwidth}
    \begin{minipage}[b]{0.45\textwidth}
        \centering
        \includegraphics[width=\textwidth]{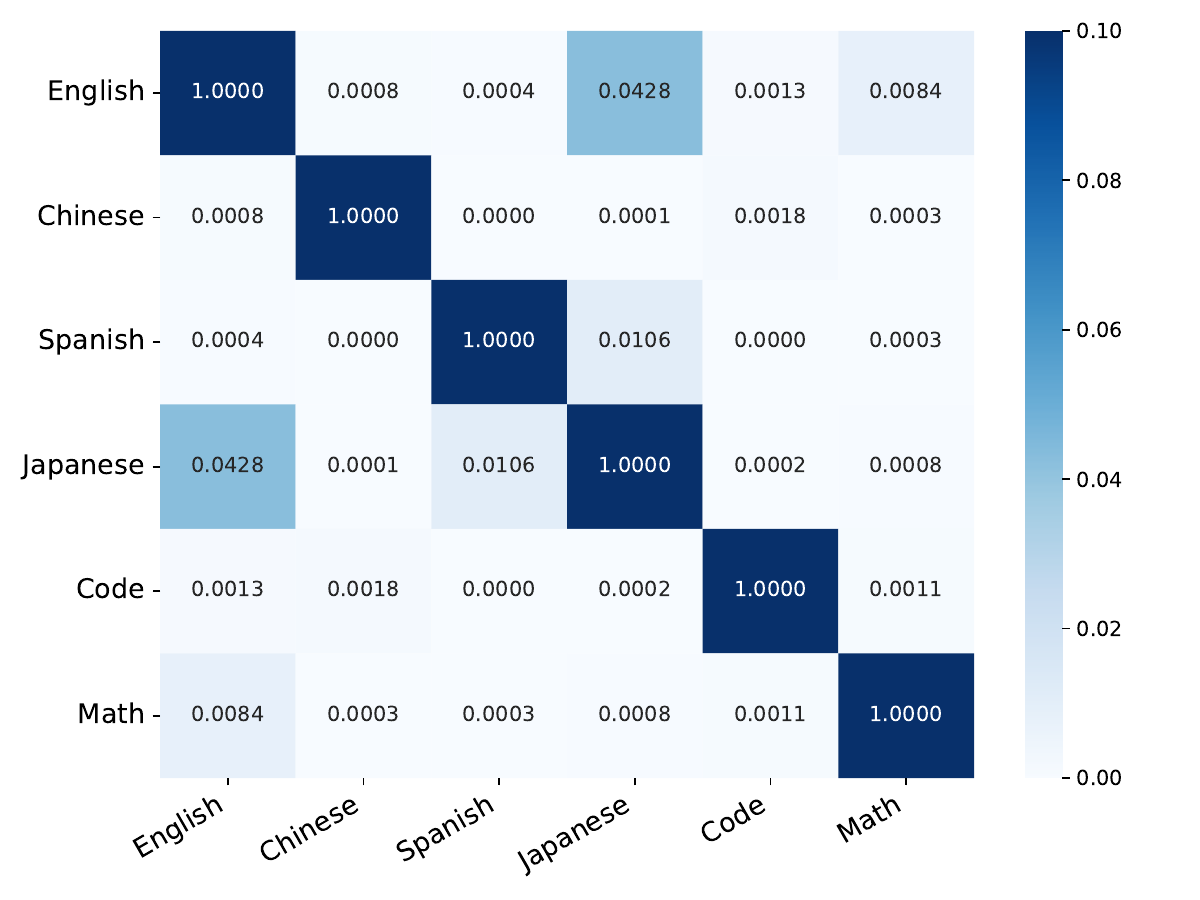}
        \caption{Verification of orthogonality. We calculate the cosine similarity between six different task vectors and find that their cosine similarity is nearly 0.}
        \label{orthogonal}
    \end{minipage}
\end{figure} 

\section{Property Verification}
In Section~\ref{method}, we introduced two properties essential to our proof.
In this section, we conduct experiments to verify these properties.

\subsection{NTK Linearization}
\citet{jacot2018neural} have proved that when the width of the neural network approaches infinity,  it demonstrates kernel behavior and the optimization proceeds in the linear regime.
We test Llama-2-7b-chat-hf~\cite{touvron2023LLaMa} on AGIEval~\cite{zhong2023agieval} dataset to verify its linearity. 
We have randomly sampled three outputs of the Llama-2-7b-chat-hf when $\alpha$ in Eq.~\ref{eq:linearization} gets value of $[0, 0.1, \cdots, 1]$.
For better visualization, we also subtract all the outputs using $\max\{y_i\}$, ensuring they have the same endpoint.
From the results in Figure~\ref{linear_fig}, we can see that all the outputs are almost linear with $\alpha$, which indicates that LLMs do exhibit a kernel behavior during finetuning.

\subsection{Task Vector Orthogonality}
\citet{ilharco2022editing, yang2023adamerging} have performed experiments to verify this property for vision models.
For LLMs, we also observe similar results: these task vectors are almost orthogonal to each other.
The result has been shown in Figure~\ref{orthogonal}. 
We can see that different task vectors are almost orthogonal, and their cosine similarity is nearly 0 as Eq.\ref{eq.orthogonal} expected, which verifies the property we have used for our proof.

\FloatBarrier

\begin{table*}[!htb]
    \centering
    \caption{ Performance comparison of merging different LLaMA-2-7B fine-tuned models on different datasets.}
    \label{LLaMa7b_compare}
    \vskip 0.1in
\resizebox{1.0\textwidth}{!}{
\setlength{\tabcolsep}{1.0mm}{
       \begin{tabular}{lcccccc|cc}
    \toprule
        Model    & WinoGrande & AGIEval & GSM8k & MATH & MBPP  & HumanEval  & Abs. Avg  & Nor. Avg    \\ \midrule 
        LM    & 62.67     & 34.01   & 28.66  & 4.00  & 22.00    & 7.31  & 26.44      & 0.91    \\
        Math  & 61.64          & 29.40   & 47.16  & 2.40 & 17.40  & 11.58  & 28.26     & 0.84     \\
        Code  & 61.88         & 27.41   & 17.21  & 2.20  & 24.80  & 21.92    & 25.90   & 0.84       \\ 
        \midrule
        
        Weight Average & 63.93   & 31.36          & 37.68 & 7.00 & 23.40&  \textbf{20.12} &  30.58   &  {1.25}   \\
        
        Task Arithmetic & 63.54 &  {31.70}           & 37.53 & 5.20 & 23.20 & 19.51   &  30.11  &   1.12  \\   
        Ties Merging   & 62.67   & 32.10     & 37.93 & 7.40                      &     22.80                     & 18.29                     & 30.20                        & 1.26     \\     
        DARE       & 63.27  & 32.25       & 37.86 & 7.00                      &    \textbf{24.40}                      & 19.51                     & 30.72                        & 1.26     \\ \rowcolor{mygray}
\textbf{\llma (ours)}    & \textbf{64.25}   & \textbf{32.71}          & \textbf{45.41} & \textbf{7.80}  & 21.20 & 17.68      & \textbf{31.51}  &  \textbf{1.31}   \\ \bottomrule           
    \end{tabular}
    }
    }
    \label{LLaMa7B}
\end{table*}
\begin{table*}[!htb]
    \centering
    \caption{Performance comparison of merging different Mistral-7B fine-tuned models on different datasets.}
    \label{Mistral7B}
    \vskip 0.1in
\resizebox{1.0\textwidth}{!}{
\setlength{\tabcolsep}{1.0mm}{
\begin{tabular}{lcccccc|cc}
    \toprule
        Model           & WinoGrande & AGIEval & GSM8k & MATH                      & MBPP  & HumanEval & Abs. Avg & Nor. Avg \\ \midrule
LM              & 69.30       & 37.55   & 47.54 & 7.80 & 34.40 & 34.75     & 38.56    &   0.776       \\
Math            & 63.46      & 38.06   & 68.46                      & 28.00                     & 24.00 & 25.00     & 41.16    &   0.854       \\
Code            & 67.32      & 40.69   & 60.73                      & 15.60                     & 43.40 & 39.02     & 44.46    &  0.917        \\ \midrule
Weight Average  & 67.88      & 41.12   & 62.77                      & 17.40                     & \textbf{40.20} & 38.41     & 44.63    &  0.921        \\
Task Arithmetic & 67.88      & 41.41   & 63.38                      & 18.80                     & \textbf{40.20} & 38.40     & 45.01    &  0.932        \\
Ties Merging    & 67.72      & 41.06   & 60.35                      & 17.80                     & \textbf{40.20} & 40.24     & 44.56    & 0.924        \\ 
DARE            & 67.40      & 40.58   & 59.67                      & 19.00                     & 36.00 & \textbf{40.85}     & 43.92    & 0.913        \\ \rowcolor{mygray}
\textbf{\llma (ours)}             & \textbf{68.35}      & \textbf{41.86}   & \textbf{66.03}                      & \textbf{20.80}                     & 39.00 & 35.37     & \textbf{45.24}    &  \textbf{0.936}        \\ \bottomrule
\end{tabular}
    }
    }
    \label{mistral}
\end{table*}

\section{Experiments}
In this section, we conduct experiments to demonstrate the effectiveness of our \llma. 
In the first section, we demonstrate that our \llma~consistently achieves optimal average performance across diverse datasets and is robust for model series with varying parameter sizes and architectures.
DARE and Ties-Merging are task vector-improving methods that resolve conflicts and redundant parameters between task vectors.
We conduct experiments to demonstrate that our method is orthogonal to theirs and can be integrated to improve the average performance further.
Finally, we show that the model merged by our \llma~has better out-of-distribution generalization ability.

\begin{table*}[htb]
\centering
\caption{Comparison of performance of merging fine-tuned LLaMA-2-13B on different datasets.}
\label{LLaMa13B}
\vskip 0.1in
\resizebox{1.0\textwidth}{!}{
\setlength{\tabcolsep}{1.0mm}{
\begin{tabular}{lcccccc|cc}
\toprule
Model           & WinoGrande & AGIEval & GSM8K & MATH & MBPP & HumanEval & Abs. Avg    & Nor. Avg    \\ \midrule
LM              & 64.80       & 35.04   & 42.84 & 4.80  & 27.00   & 15.24     & 31.62       & 1.02 \\
Math             & 60.38      & 36.74   & 55.27 & 3.40  & 22.60 & 12.80      & 31.87      & 0.93 \\
Code          & 63.93      & 32.04   & 36.47 & 5.00    & 26.60 & 16.46     & 30.08 & 1.01 \\ \midrule
Weight Average  & 64.88      & 37.23   & \textbf{53.15} & 7.60  & 29.80 & 21.95     & 35.77 & 1.29 \\
Task Arithmetic & 65.11      & 35.48   & 50.34 & 7.20  & 29.80 & 21.95     & 34.98       & 1.25 \\
Ties Merging    & 65.23          & 36.02&51.23&7.40&30.20&\textbf{23.17}&35.54&1.28    \\
DAREs     & \textbf{65.70} &36.87&51.85&7.60&30.00&22.56&35.76&1.29    \\ \rowcolor{mygray}
\textbf{\llma (ours)}            & 65.04      & \textbf{37.33}   & 52.92 & \textbf{7.80}  & \textbf{30.40} & 21.95     & \textbf{35.91} & \textbf{1.30} \\ \bottomrule
\end{tabular}
        }
    }
\end{table*}
\begin{table*}[htb]
\caption{\llma~can be integrated with DARE and Ties-Merging, thereby leading to further improvment.}
\label{integrate}
\vskip 0.1in
\resizebox{1.0\textwidth}{!}{
\setlength{\tabcolsep}{1.0mm}{
\begin{tabular}{lcccccc|cc}
\toprule
Method     & WinoGrande & AGIEval & GSM8k & MATH & MBPP & HumanEval & Abs. Avg    & Nor. Avg    \\ \midrule
Ties-Merging & 62.67      & 32.10   & 37.93 & 7.40  & 22.80 & 18.29     & 30.20 & 1.26 \\ \rowcolor{mygray}
\textbf{Ties + \llma}  & 62.35      & 32.91   & 46.10 & 8.00    & 22.40 & 17.68     & \textbf{31.57} & \textbf{1.33} \\ \midrule
Dare       & 63.27      & 32.25   & 37.86 & 7.00    & 24.40 & 19.51     & 30.72      & 1.26 \\ \rowcolor{mygray}
\textbf{Dare + \llma}  & 62.99      & 33.01   & 45.72 & 7.60  & 21.80 & 18.29     & \textbf{31.57} & \textbf{1.30} \\ \bottomrule        
\end{tabular}
}
}
\end{table*}

\subsection{Merging Models Using \llma}
\paragraph{Dataset and Models.}

To test the effectiveness of our method, we use Llama-2-7b-chat-hf~\cite{touvron2023LLaMa}, MAmmoTH-7B~\cite{yue2023mammoth} and llama-2-coder-7b~\cite{manuel_romero_2023} as models fine-tuned on general knowledge, math, and code datasets using the pre-trained model Llama-2-7B-hf~~\cite{touvron2023LLaMa}.
Moreover, we use a different model architecture: Mistral-7B-Instruct-v0.2~\cite{Mistral-7B-Instruct}, MAmmoTH2-7B-Plus~\cite{yue2024mammoth2} and Mistral-7B-codealpaca-lora~\cite{Nondzu} as models fine-tuned on general knowledge, math, and code datasets using pre-trained model Mistral 7B~\cite{jiang2023mistral}.
We also provide experiments using models with larger sizes: Llama-2-13b-chat-hf~\cite{touvron2023LLaMa}, MAmmoTH-13B~\cite{yue2023mammoth}, and llama-2-13b-code-chat~\cite{LLaMa-2-13b-code-chat} as models fine-tuned on general knowledge, math, and code datasets using the pre-trained model Llama-2-13B-hf~\cite{touvron2023LLaMa}.
We use WinoGrande~\cite{sakaguchi2021winogrande} and AGIEval~\cite{zhong2023agieval} for evaluating general knowledge performance, GSM8K~\cite{cobbe2021gsm8k} and MATH~\cite{2019arXiv} for testing mathematical reasoning ability, HumanEval~\cite{chen2021evaluating} and MBPP~\cite{austin2021program} for estimating code-generation capacity.

\paragraph{Evaluation Metrics.}
We use common evaluation settings for a single task: 5-shot accuracy for AGIEval, 4-shot accuracy for GSM8K and MATH, 3-shot accuracy for MBPP, and zero-shot accuracy for HumanEval and WinoGrande.
We employ two key metrics in evaluating different merging methods: absolute average performance and normalized average accuracy.

\paragraph{Quantitative Evaluation for LLaMA-2-7B.}
We use the metrics and datasets we introduced above to evaluate the performance of different methods.
We use Weight Average~\cite{wortsman2022model}, Task Arithmetic~\cite{ilharco2022editing}, Ties-Merging~\cite{yadav2024ties} and DARE~\cite{yu2023language}, which are also model exclusive and computationally efficient methods, to compare with our method by merging LLaMA-2-7B.
The scores in Table~\ref{LLaMa7b_compare} show that for WinoGrande, AGIEval, GSM8k, and MATH dataset, our method scores 64.25, 32.71, 45.41, and 7.80, which outperforms other methods. For the HumanEval dataset, DARE performs best, and for the MBPP dataset, the Weight Average method achieves the highest score.
Since our method aims to achieve the \textit{average best performance}, we use absolute average performance score and normalized average performance score to compare the five methods. 
We can see that our \llma~achieves the rank-1 score 31.51, 1.31 in both absolute average performance and normalized average performance.

\paragraph{Using Different Model Architecture.}
We also use a different model architecture, Mistral-7B, for evaluation, and the result has been shown in Table~\ref{Mistral7B}.
The scores in Table~\ref{Mistral7B} show similar results to LLaMA-2-7B: For WinoGrande, AGIEval, GSM8k, and MATH dataset, our \llma~scores 41.86, 68.35, 66.03, 20.8, which outperforms existing methods, for HumanEval dataset Weight Average, Task Arithmetic, and Ties Merging performs best and for MBPP dataset, DARE method achieves the highest score.

\paragraph{Using Larger Model Size.}

We also test our method using a larger model LLaMA-2-13B~\cite{touvron2023LLaMa}. 
The scores in Table~\ref{LLaMa13B} demonstrate that for AGIEval, Math, and MBPP datasets, our method outperforms other methods.
For WinoGrand, GSM8K, and HumanEval dataset, DARE, Weight Average and Ties-Merging achieves the highest score.
Similarly, under the \textit{average measure} absolute average performance and normalized average performance, our method also outperforms the other five methods.

\paragraph{Integrate with Ties/DARE}

As there are conflicts and redundant parameters between task vectors,
DARE~\cite{yu2023language} and Ties-Merging~\cite{yadav2024ties} are two methods trying to solve the interfaces, reducing the redundancy and thereby improving the performance of task arithmetic. 
Since our method is also based on the framework of task arithmetic, Ties-merging and DARE are expected to improve the performance of our \llma~ further.
As we can see in Table~\ref{integrate}, under the baseline of Ties-Merging and DARE methods, our method is orthogonal to Ties-Merging and DARE and can integrate them into our \llma, thus leading to further improvement.
For example, the average absolute performance of DARE has been improved by our \llma~from 30.72 to 31.57. And the normalized absolute performance of DARE has been improved by our \llma~from 1.26 to 1.3. 
Ties-merging also leads to a similar conclusion: the average absolute performance of DARE has been improved by our \llma~from 30.20  to 31.57. And the normalized absolute performance of DARE has been improved by our \llma~from 1.26 to 1.33.

\subsection{Out of Distribution Generalization}
Following~\cite{yang2023adamerging, jin2022dataless}, we also compare the out-of-distribution generalization ability of different merging methods.
We evaluate different methods using JEC-QA~\cite{zhong2020jec}, FinanceIQ~\cite{financeiq}, and MedQA~\cite{jin2021disease} dataset. All three datasets use 5-shot accuracy as the evaluation metric.
Table~\ref{OODtab} summarizes out-of-distribution generalization performance when merging all domain specific models using different methods.
As we can see, \llma~outperforms current methods on these unseen datasets, which demonstrates that \llma~is more robust to the test data distribution shifts.

\begin{table}[htb]
\centering
\caption{Out of Distribution Generalization}
\label{OODtab}
\hspace{-.3cm}
\resizebox{1.\textwidth}{!}{
\setlength{\tabcolsep}{8.0mm}{
\begin{tabular}{lccc|c}
\toprule
Model           & JEC-QA & FinancelQ & MedQA & Avg \\ \midrule
LM              & 31.32 & 32.83     & 30.20 & 31.45   \\
Math            & 25.56 & 30.25     & 24.73 & 26.85   \\
Code            & 29.23 & 30.87     & 26.25 & 28.78   \\ \midrule
Weight Average  & 30.73 & 34.17     & 29.90 & 31.60   \\
Task Arithmetic & 30.85 & 33.89     & 30.13 & 31.62   \\
Ties Merging            & 30.80 & 33.53     & 30.02 & 31.45   \\
DARE            & 30.79 & 33.93     & \textbf{30.17} & 31.63   \\
\rowcolor{mygray}
\textbf{\llma (ours)} & \textbf{30.97} & \textbf{34.31}     & 30.07 & \textbf{31.78}   \\ \bottomrule
\end{tabular}
}
}
\end{table}

\section{Conclusion}
In this paper, we have provided a novel model merging method named \llma, an efficient and optimal model-exclusive task arithmetic specifically designed for LLMs.
We provide the mathematical formulation of task arithmetic's optimization objective and the theoretical analysis of the task arithmetic performance bound.
By separating the data and scaling coefficient term under careful approximation, the closed-form solution provides an avenue for optimally achieving task arithmetic without using any data.
Extensive experiment results show that our \llma~outperforms the existing state-of-the-art model-exclusive merging method and can be integrated with task vector-improving methods such as Ties-Merging and DARE.

\section{Limitations}
(1) Our works share the same general limitation of existing task arithmetic based methods: Our merging method relies on common initialization and model architecture, which ensures that the task vectors are orthogonal.
(2) Moreover, since our method is specifically designed for LLMs and relies on the NTK linearization, for small size models, our method may not perform well.
%


\bibliography{baichuan}
\bibliographystyle{baichuan}

{\onecolumn
\newpage}
\appendix
\section*{Appendix}
\section{Proof}
\label{sec:appendix}
\subsection{Proof of Lemma~\ref{LLema4}}
Using Taylor expansion for $\mathcal{L}(\vth_{\textup{final}}, \bm{x})$ at $\vth_0$:
\begin{align}
    \mathcal{L}(\vth_{\textup{final}}, \bm{x}) 
    =& \mathcal{L}_t(\sum_{k=1}^n\lambda_k(\vth_{k} - \vth_0)+\vth_0, \vx_t) \\
    =& \mathcal{L}_{t}(\bm{h}_t + \vth_t  , \vx_t) \\ \notag
    =& \mathcal{L}_{t}(\vth_t  , \vx_t) + \nabla \mathcal{L}_{t}(\vth_t  , \vx_t)\bm{h}_t
    + \frac{1}{2}\bm{h}_t^\top\left(\int_{0}^{1}\nabla^2 \mathcal{L}_t(\gamma_t(\beta))d\beta \right)\bm{h}_t
    \label{19}
\end{align}
where $\gamma_t(\beta) = \vth_t+\beta(\vth_{\text{final}} - \vth_t)$ and $\bm{h}_t$ is the linear combination of $\bm{\lambda}$ and $\vth$:
\begin{align}
    \bm{h}_t = \sum_{k\neq t} \lambda_k(\vth_k - \vth_0) - (1- \lambda_{t})(\vth_{t} - \vth_0)
\end{align}
Because the $\vth_t$ is fine-tuned using loss $\mathcal{L}_t$, the gradient of $\mathcal{L}_t$ at $\vth_t$ is zero, and the first order expansion is 0.
Substituting Eq.~\ref{19} to Eq.~\ref{tldeq}, we have:
\begin{align}
    \tldt 
    &= \mathcal{L}_t(\vth_{\textup{final}}, \vx_t) - \mathcal{L}_t(\vth_{t}, \vx_t)\\
    &= \frac{1}{2}\bm{h}_t^\top\left(\int_{0}^{1}\nabla^2 \mathcal{L}_t(\gamma_t(\beta))d\beta \right)\bm{h}_t
\end{align}
Thus, we have completed the proof.
\subsection{Proof of Theorem~\ref{thm7}}
Before starting the proof, we first introduce a lemma:
\begin{lemma}
\label{Lemma11}
    Under the Property.~\ref{linear}, the task vector is linearly with the gradient.
    \begin{align}
         \delta_t(\vth_t - \vth_0) =  \nabla_{\vth_0} f(\vx, \vth_0)
    \end{align}
\end{lemma}
\noindent\textbf{Proof:}
For gradient descent, we have:
\begin{align}
        \vth_t - \vth_0 
        &= \sum_{i=1}^n lr_i \nabla \mathcal{L}_t^i
        = \sum_{i=1}^n lr_i \frac{\partial \mathcal{L}_t^i}{\partial f}\nabla {f}_i
\end{align}
where $lr_i$ and $\nabla \mathcal{L}_t^i$ and $\nabla f_i$ is the learning rate, gradient loss, gradient of $f$ at step i.
From Property~\ref{linear}, we can see that the fine-tuning process of $f$ occurs in the linear regime, which indicates that the first order derivative in the task vector direction is an constant. We derivative at $\vth_t$:
\begin{align}
    \nabla_{\vth_t} f(\vx, \vth_t) = \nabla_{\vth_0} f(\vx, \vth_0)
\end{align}
Thus, we substitute all the gradient of $f_i$ using  $\nabla_{\vth_0} f(\vx, \vth_0)$:
\begin{align}
      \delta_t(\vth_t - \vth_0) = \nabla_{\vth_0} f(\vx, \vth_0)
\end{align}
$$\text{where}\quad \frac{1}{\delta_t} = \sum_{i=1}^n lr_i \frac{\partial \mathcal{L}_t^i}{\partial f}$$

Thus, we have completed the proof of the Lemma.

For the of loss function, using Property~\ref{linear} we have:
\begin{align}
\mathcal{L}_t(\vth_t, \vx_t)
=& \frac{1}{2}\left\Vert f(\bm{x}_t, \vth_t) - y\right\Vert^2  
= \frac{1}{2} \Vert (\vth_t - \vth_0)^\top\nabla f(\vx_t;\vth_0) + C_0 \Vert^2
\end{align}
For the Hessian of loss function, it can be represented as:
\begin{align}
\nabla^2_{\vth_t} \mathcal{L}_t&= \nabla_{\vth_0}f(\vx_t;\vth_0)\nabla^\top_{\vth_0}f(\vx_t;\vth_0)
\label{26}
\end{align}
Using Eq.~\ref{26} the $\tldt$ can be represented as:
\begin{align}
    2 \tldt 
    =&\bm{h}_t^\top\left(\int_{0}^{1}\nabla^2 \mathcal{L}_t(\gamma_t(\beta))d\beta \right)\bm{h}_t \\
    =&  \bm{h}_t^\top\left(\nabla^2 \mathcal{L}_t(\Tilde{\vth})\right) \bm{h}_t  \\
    =&  \bm{h}_t^\top\left(\nabla_{\vth_0} f(\vth_0, \bm{x}_t) \nabla_{\vth_0}  f^\top(\vth_0, \bm{x}_t)\right)\bm{h}_t  \\
    =& \tr\bigg\{\bm{h}_t^\top\left(\nabla_{\vth_0}  f(\vth_0, \bm{x}_t) \nabla_{\vth_0}  f^\top(\vth_0, \bm{x}_t)\right)\bm{h}_t\bigg\}  \\
    \leq&  \tr(\vh\vh^\top)\tr\left(\nabla_{\vth_0}  f(\vth_0, \bm{x}_t) \nabla_{\vth_0} f^\top(\vth_0, \bm{x}_t)\right)
\end{align}
For $\tr(\vh\vh^\top)$, using Property.~\ref{orthogonality}, we have:
\begin{align}
     \tr(\vh\vh^\top)
     =&  \left\Vert \sum_{k\neq t}^T \lambda_k(\vth_k - \vth_0) - (1- \lambda_{t})(\vth_{t} - \vth_0)^\top \right\Vert^2\\
     =& \sum_{k\neq t}^T
        \left[
        \vmathbb{1}_{k\neq t}(\lambda^2_k)
        +\vmathbb{1}_{k= t}(1 - \lambda^2_k)\right]
        \Vert \vth_k -  \vth_0\Vert^2 \\
    =& \sum_{k\neq t}^T
        \left[
        \vmathbb{1}(\lambda^2_k)
        \Vert \vth_k -  \vth_0\Vert^2 \right]
\end{align}
$$\textup{where}\quad (\lambda^2_k)\vmathbb{1}(k\neq t) +(1 - \lambda^2_k)\vmathbb{1}(k= t) := \vmathbb{1}_t(\lambda_k^2).$$
For the second part: $\tr(\nabla_{\vth_0}  f(\vth_0, \bm{x}_t) \nabla_{\vth_0}  f^\top(\vth_0, \bm{x}_t))$, using Lemma~\ref{Lemma11} we can have:
\begin{align}
    \tr\left(\nabla_{\vth_0}  f(\vth_0, \bm{x}_t) \nabla_{\vth_0}  f^\top(\vth_0, \bm{x}_t)\right)
    =\delta_t^2\Vert \vth_t - \vth_0\Vert^2
\end{align}
Thus, for $\tldt$ we can upper bound it by
\begin{align}
    \tldt &\leq \frac{\delta_t^2}{2}
          \left\Vert \vth_t - \vth_0  \right\Vert_2^2 \bigg\{ \sum_{k\neq t}^T
        \vmathbb{1}_t(\lambda_k^2)
        \Vert \vth_k -  \vth_0\Vert^2 \bigg\} 
        \label{42}
\end{align}
\subsection{Proof of Theorem~\ref{thm8}}
By summing Eq.\ref{42} from 1 to T, we can complete the proof.
\begin{align}
    \textup{ALD} &\leq \sum_{t=1}^T
         \frac{\delta_t^2}{2}\left\Vert \vth_t - \vth_0  \right\Vert_2^2 \bigg\{ \sum_{k\neq t}^T
        \vmathbb{1}_t(\lambda_k^2)
        \Vert \vth_k -  \vth_0\Vert^2 \bigg\} 
        \label{43}
\end{align}
\subsection{Proof of Theorem~\ref{thm9}}
First, for Eq.~\ref{43}, we have:
\begin{align}
    \textup{ALD} &\leq\frac{\delta_t^2}{2}  \sum_{t=1}^T
         \left\Vert \vth_t - \vth_0  \right\Vert_2^2 \bigg\{ \sum_{k\neq t}^T
        \vmathbb{1}_t(\lambda_k^2)
        \Vert \vth_k -  \vth_0\Vert^2 \bigg\} 
        \label{44}
\end{align}
where $\delta_0 = \max\{\delta_i\}$ For Eq.~\ref{44}, it is easy to verify that the terms containing $\lambda_t$ can be represented as:
\begin{align}
    \textup{ALD}_{\lambda_t} =& \frac{\delta_t^2}{2} \Vert \vth_t  -\vth_0\Vert^2 \left[\sum_{k=1}^T \vmathbb{1}_t(\lambda) \Vert \vth_k - \vth_0\Vert^2\right]
\end{align}
Thus, the ALD can be upper bounded by
\begin{align}
        \textup{ALD}\leq 
        \sum_{t=1}^T \textup{ALD}_{\lambda_t} 
\end{align}
\subsection{Proof of Theorem~\ref{thm10}}
Because each $\text{ALD}_{\lambda_t}$ does not contain other scaling coefficients.
We can solve each optimal $\lambda_t$ from $\text{ALD}_{\lambda_t}$:
    \begin{align}
        \lambda_t 
        &= \mathop{\arg\min}_{\lambda_t}\frac{\delta_0^2}{2} \Vert \vth_t - \vth_0 \Vert^2\left[\sum_{k=1}^T \vmathbb{1}_t(\lambda) \Vert \vth_k - \vth_0\Vert^2\right]\\
        &= \mathop{\arg\min}_{\lambda_t} \Vert \vth_t - \vth_0 \Vert^2\left[\sum_{k=1}^T \vmathbb{1}_t(\lambda) \Vert \vth_k - \vth_0\Vert^2\right]
    \end{align}
The RHS of the above equation is quadratic on $\lambda_t$ and and the optimal solution for $\lambda_t$ is:
    \begin{align}
        \lambda_t = \frac{\Vert \vth_t - \vth_0\Vert^2}{\sum_{k=1}^n\Vert \vth_k - \vth_0\Vert^2}
    \end{align}
\section{Details of Models and Datasets}
Table~\ref{Dataset Details} shows the versions and correspondence with pre-trained backbones of fine-tuned LLMs.
Table~\ref{Model Details} shows the details of the datasets we use in our paper.

\section{Infra and hardware details}
We use PyTorch as the deep learning framework. We merge and evaluate the neural networks using A100 GPUs. 
\section{Hyper-parameter Setting}
For both DARE and TIES-Merging, the density of 0.55 is used, and the open-source tool MergeKit\footnote{\href{https://github.com/arcee-ai/mergekit/tree/main}{MergeKit}} is employed for the merging process.

\section{Details of different Methods}\
We give a detailed comparison of the current merging method below from the perspective of extra data information, time complexity, and optimal performance.
The time complexity for forward and backward processes is denoted as FW and BP.
For RegMean, it requires the inner product data matrices for layer input to calculate the updated parameters. It only requires a forward process, but loading all the inner products of the layer input matrix requires $\mathcal{O}(\theta^2)$ memory.
For Fisher merge, it also requires the data to calculate the Fisher Matrix, which requires the forward process to calculate the Fisher matrix and $\mathcal{O}(\theta^2)$ memory to store the Fisher matrix.
Grid-search Task Arithmetic (G-Task Arithmetic) requires $\mathcal{O}(G\textsuperscript{T}\times\mathcal{T}_{\text{FW}})$ forward process to evaluate, where G is the grid number (G = 100 means 100 girds from 0 to 1) and T is the number of tasks.
The space complexity is also equal to the memory requirement of the forward process.
For Adamerging, it simultaneously loads T LLMs to optimize, whose time complexity is  $\mathcal{O}(\mathcal{T}_{\text{BP}})$ and space complexity is:  $\mathcal{O}(\mathcal{S}_{\text{BP}}\times T)$.
For weight average, task arithmetic, and \llma, they all do not need extra data information, which is model exclusive. 
Their time and space complexity is $\mathcal{O}(1)$ and $\mathcal{O}(n)$, but only our \llma~achieves optimal performance.

\begin{table*}[htb]
\centering
 \caption{Extra data information requirement, time and space complexity, and optimally of current methods. The time complexity for forward process and back propagation are denote by $\mathcal{T}_{\text{FW}}, \mathcal{T}_{\text{BP}}$. The space complexity for forward process and back propagation are denote by $\mathcal{S}_{\text{FW}}, \mathcal{S}_{\text{BP}}$. T is the number of task, $\theta$ is the number of parameters and G is the grid number (G = 100 means 100 girds from 0 to 1).}
 \label{complexity}
 \vskip 0.1in
 \resizebox{1.\textwidth}{!}{
 \setlength{\tabcolsep}{1.0mm}{
 \begin{tabular}{lccccc}
 \toprule
                   & Extra Data Info & Time Complexity & Space Complexity & Optimal & Apply to LLMs\\ \midrule
 RegMean           &  \Checkmark  & $\mathcal{O}(\mathcal{T}_{\text{FW}})$   & $\mathcal{O}(\theta^2)$       & \Checkmark   & \XSolidBrush  \\
Fisher Merge      &  \Checkmark & $\mathcal{O}(\mathcal{T}_{\text{FW}})$   & $\mathcal{O}(\theta^2)$   & \Checkmark   & \XSolidBrush\\
 G-Task Arithmetic &  \Checkmark &  $\mathcal{O}(G\textsuperscript{T}\times\mathcal{T}_{\text{FW}})$ & $\mathcal{O}(\mathcal{S}_{\text{FW}})$  & \Checkmark   & \XSolidBrush\\
 AdaMerging        & \Checkmark & $\mathcal{O}(\mathcal{T}_{\text{BP}})$     & $\mathcal{O}(\mathcal{S}_{\text{BP}}\times T)$  &  \Checkmark & \XSolidBrush\\
 Task Arithmetic & \XSolidBrush    &  $\mathcal{O}(1)$  & $\mathcal{O}(\theta)$     & \XSolidBrush &  \Checkmark \\
 Weight Average & \XSolidBrush    &  $\mathcal{O}(1)$   & $\mathcal{O}(\theta)$    & \XSolidBrush  &  \Checkmark\\
 \llma             & \XSolidBrush  & $\mathcal{O}(1)$  & $\mathcal{O}(\theta)$     & \Checkmark   &  \Checkmark \\ \bottomrule
 \end{tabular}
 }
 }
 \end{table*}

\begin{table*}[htb]
\caption{Details of datasets we used for our evaluation.}
\label{Dataset Details}
\centering
 \vskip 0.1in
 \resizebox{1.\textwidth}{!}{
 \setlength{\tabcolsep}{1.0mm}{
\begin{tabular}{lcccc}
\toprule
Dataset     & Number of Training Examples & Number of Validation Examples & Number of Testing Examples  & Evaluate Metric\\ \midrule
WinoGrande   & 9,248      & 1,267   & 1,767   & 0-shot accuracy    \\ \midrule
AGIEval& N/A      & N/A   &8,062 &  5-shot accuracy \\ \midrule
GSM8k& 7,473      & N/A   & 1,319   & 4-shot accuracy    \\ \midrule
Math& 7,500     & N/A   & 1,500    & 4-shot accuracy    \\ \midrule
MBPP& 374      &30   & 500    & 3-shot accuracy    \\ \midrule
HumanEval& N/A     & N/A   & 164    & 0-shot accuracy   \\ \midrule
JEC-QA   & N/A      & N/A   & 26,365   & 5-shot accuracy   \\ \midrule
FinancelQ& N/A      & N/A   & 7,173   & 5-shot accuracy    \\ \midrule
MedQA & N/A      & N/A   & 61,097     & 5-shot accuracy \\ 
\bottomrule
\end{tabular}
}
}
\end{table*}
\begin{table*}[htb]
\centering
\caption{Details of models we used for our evaluation.}
\label{Model Details}
\begin{tabular}{lll}
\toprule
Pre-trained Model            & Task                   & Fine-tuned-Models                   \\ \midrule
\multirow{6}{*}{LLaMA-2-7b}  & General Knowledge      & meta-llama/Llama-2-7b-chat-hf      \\
                             & Mathematical Reasoning & TIGER-Lab/MAmmoTH-7B               \\
                             & Code Generating        & mrm8488/llama-2-coder-7b           \\
                             & Chinese                & hfl/chinese-llama-2-7b             \\
                             & Spanish                & clibrain/Llama-2-7b-ft-instruct-es \\
                             & Japanese               & elyza/ELYZA-japanese-Llama-2-7b    \\ \midrule
\multirow{3}{*}{Mistral-7b}  & General Knowledge      & mistralai/Mistral-7B-Instruct-v0.2 \\
                             & Mathematical Reasoning & TIGER-Lab/MAmmoTH2-7B              \\
                             & Code Generating        & Nondzu/Mistral-7B-codealpaca-lora  \\ \midrule
\multirow{3}{*}{LLaMA-2-13b} & General Knowledge      & meta-llama/Llama-2-13b-chat-hf     \\
                             & Mathematical Reasoning & TIGER-Lab/MAmmoTH-13B              \\
                             & Code Generating        & emre/llama-2-13b-code-chat         \\ \bottomrule
                             
\end{tabular}
\end{table*}

\FloatBarrier

\section{Merging Checkpoints of the Pre-trained Model }\

We further conducted experiments using intermediate checkpoints from the open-source model Baichuan2-7B~\cite{baichuan2}. Surprisingly, we found that applying our \llma  to pre-trained model also lead to performance improvements across multiple tasks. We tested our approach on MMLU, CMMLU, GSM8k, MATH, and MBPP evaluations. The results showed that the merged model achieved an average score of 29.376, which surpasses the best individual checkpoint score of 29.334 and the weight average score of 27.800. Moreover, our approach consistently outperformed the baseline method of Weight Average. These findings underscore the robustness and effectiveness of our methodology in enhancing model performance beyond standard averaging techniques.

\begin{table*}[htb]
\caption{Performance of different checkpoints.}
\label{Checkpoints_Details}
\centering
\hspace{-.3cm}
\resizebox{1.\textwidth}{!}{
\setlength{\tabcolsep}{4.0mm}{
\begin{tabular}{lcccccc}
\toprule
\# of Tokens & MMLU & CMMLU & GSM8k & MATH & MBPP & Avg \\
\midrule
1700B   & 49.95   & 51.17   & 16.15   & 4.0     & 18.0    & 27.854  \\
1710B   & 48.67   & 51.61   & 16.53   & 3.8     & 17.8    & 27.682  \\
1720B   & 49.85   & 51.37   & 17.06   & 2.8     & 18.8    & 27.976  \\
1730B   & 49.91   & 50.24   & 17.44   & 4.2     & 19.2    & 28.198  \\
1740B   & 50.99   & 49.64   & 19.64   & 3.8     & 22.6    & 29.334  \\
1750B   & 49.50   & 51.20   & 17.13   & 4.6     & 19.2    & 28.326  \\
1760B   & 49.02   & 48.52   & 17.29   & 4.2     & 17.8    & 27.366  \\
1770B   & 51.48   & 49.74   & 17.36   & 4.4     & 20.4    & 28.676  \\
\midrule
Weight Average     & 48.95   & 50.01   & 17.44   & 4.4     & 18.2    & 27.800  \\
\textbf{\llma (ours)}    & 50.61   & \textbf{52.69}   & 19.18   & 4.4     & 20.0     & \textbf{29.376}  \\
\bottomrule
\end{tabular}
}
}
\end{table*}

\end{document}